\documentclass[twoside,11pt]{article}
\usepackage[utf8]{inputenc}
\usepackage{jair, theapa, rawfonts}

\usepackage[nolist]{acronym}
\usepackage{graphicx}
\usepackage{booktabs}
\usepackage{multirow}
\usepackage{arydshln}
\usepackage{xcolor}
\usepackage{amsmath}

\newcommand{\todo}[1]{} 
\renewcommand{\todo}[1]{{\noindent \color{red} \textbf{TODO: } {#1}}}

\DeclareMathOperator*{\argmax}{argmax}

\ShortHeadings{General-Purpose Communicative Function Recognition}
{Ribeiro, Ribeiro, \& Martins de Matos}
\firstpageno{1}

\begin{document}

\title{Automatic Recognition of the General-Purpose Communicative Functions defined by the ISO 24617-2 Standard for Dialog Act Annotation}

\author{\name Eug\'{e}nio Ribeiro \email eugenio.ribeiro@inesc-id.pt \\
    \addr INESC-ID Lisboa, Portugal \\
    Instituto Superior T\'{e}cnico, Universidade de Lisboa, Portugal
    \AND
    \name Ricardo Ribeiro \email ricardo.ribeiro@inesc-id.pt \\
    \addr INESC-ID Lisboa, Portugal \\
    Instituto Universit\'{a}rio de Lisboa (ISCTE-IUL), Portugal
    \AND
    \name David Martins de Matos \email david.matos@inesc-id.pt \\
    \addr INESC-ID Lisboa, Portugal \\
    Instituto Superior T\'{e}cnico, Universidade de Lisboa, Portugal
}


\maketitle

%

\begin{acronym}
    \acro{ASR}{Automatic Speech Recognition}
    \acro{CNN}{Convolutional Neural Network}
    \acro{CRF}{Conditional Random Field}
    \acro{DNN}{Deep Neural Network}
    \acro{FCT}{Fundação para a Ciência e a Tecnologia}
    \acro{GRU}{Gated Recurrent Unit}
    \acro{LSTM}{Long Short-Term Memory Unit}
    \acro{MAP}{Maximum a Posteriori}
    \acro{NLP}{Natural Language Processing}
    \acro{POS}{Part-of-Speech}
    \acro{QA}{Question Answering}
    \acro{ReLU}{Rectified Linear Unit}
    \acro{RNN}{Recurrent Neural Network}
    \acro{SVM}{Support Vector Machine}
    \acro{WER}{Word Error Rate}
\end{acronym}

%
%

\begin{abstract}

ISO 24617-2, the standard for dialog act annotation, defines a hierarchically organized set of general-purpose communicative functions. The automatic recognition of these functions, although practically unexplored, is relevant for a dialog system, since they provide cues regarding the intention behind the segments and how they should be interpreted. We explore the recognition of general-purpose communicative functions in the DialogBank, which is a reference set of dialogs annotated according to this standard. To do so, we propose adaptations of existing approaches to flat dialog act recognition that allow them to deal with the hierarchical classification problem. More specifically, we propose the use of a hierarchical network with cascading outputs and maximum a posteriori path estimation to predict the communicative function at each level of the hierarchy, preserve the dependencies between the functions in the path, and decide at which level to stop. Furthermore, since the amount of dialogs in the DialogBank is reduced, we rely on transfer learning processes to reduce overfitting and improve performance. The results of our experiments show that the hierarchical approach outperforms a flat one and that each of its components plays an important role towards the recognition of general-purpose communicative functions. 

\end{abstract}

%

%

\section{Introduction}
\label{sec:introduction}

From the perspective of a dialog system, it is important to identify the intention behind the segments in a dialog, since it provides an important cue regarding the information that is present in the segments and how they should be interpreted. According to \citeA{Searle1969}, that intention behind the uttered words is revealed by the corresponding dialog acts, which he defines as the minimal units of linguistic communication. Consequently, automatic dialog act recognition is an important task in the context of \ac{NLP}, which has been widely explored over the years. In an attempt to set the ground for more comparable research in the area, \citeA{Bunt2012} defined the ISO 24617-2 standard for dialog act annotation. However, annotating dialogs according to this standard is an exhaustive process, especially since the annotation of each segment does not consist of a single dialog act label, which in the standard nomenclature is called a communicative function, but rather of a complex structure which includes information regarding the semantic dimension of the dialog acts and relations with other segments, among other aspects. Consequently, the amount of data annotated according to the standard is still small and the automatic recognition of its communicative functions remains practically unexplored. 

In this article, we explore the automatic recognition of communicative functions in the English dialogs available in the DialogBank~\cite{Bunt2016,Bunt2019}, which, to the best of our knowledge, is the only publicly available source of dialogs fully annotated according to the standard. We focus on general-purpose communicative functions, since they are predominant and, contrarily to the dialog act labels of widely explored corpora in dialog act recognition research, they pose a hierarchical classification problem, with paths that may not end on a leaf communicative function.


To approach the problem, we propose modifications to existing approaches on dialog act recognition that allow them to deal with the hierarchical classification problem posed by the general-purpose communicative functions defined by the ISO 25617-2 standard. These modifications focus on the ability to predict communicative functions at the multiple levels of the hierarchy, identify when the available information is not enough to predict more specific functions, and preserve the dependencies between the functions in the path. Furthermore, given the reduced amount of annotated dialogs provided by the DialogBank, we explore the use of additional dialogs annotated using mapping processes. These processes do not lead to a complete annotation according to the standard, but the dialogs can still provide relevant information for training automatic approaches to communicative function recognition. Additionally, we rely on pre-trained dialog act recognition models by using them in transfer learning processes. This way, we can take advantage of their ability to capture generic intention information and focus on identifying that which is most relevant for recognizing the general-purpose communicative functions defined by the standard.

In the remainder of the article, we start by providing an overview on the ISO 25617-2 standard and on dialog act recognition approaches in Section~\ref{sec:related}. Then, in Section~\ref{sec:approach}, we describe our approach for predicting the general-purpose communicative functions of the standard. Section~\ref{sec:setup} describes our experimental setup, including the datasets, evaluation methodology, and implementation details. The results of our experiments are presented and discussed in Section~\ref{sec:results}. Finally, Section~\ref{sec:conclusion} summarizes the contributions of the article and provides pointers for future work.

%
%

\section{Related Work}
\label{sec:related}

Since we are exploring the automatic recognition of the communicative functions defined by the standard for dialog act annotation, in this section we start by providing an overview on that standard. Then, we discuss state-of-the-art approaches on dialog act recognition and the few which have been applied to communicative function recognition.

\subsection{ISO Standard for Dialog Act Annotation}
\label{sec:related:standard}

ISO 24617-2, the ISO standard for dialog act annotation~\cite{Bunt2012,Bunt2017} aims at setting the ground for more comparable research in the area. According to it, dialog act annotations should not be performed on turns or utterances, but rather on functional segments~\cite{Carroll1978}. Furthermore, the annotation of each segment does not consist of a single label or set of labels, but rather of a complex structure containing information about the participants, relations with other functional segments, the semantic dimension of the dialog act, its communicative function, and optional qualifiers concerning certainty, conditionality, partiality, and sentiment. 

\begin{figure}
    \centering
    \includegraphics[width=\textwidth]{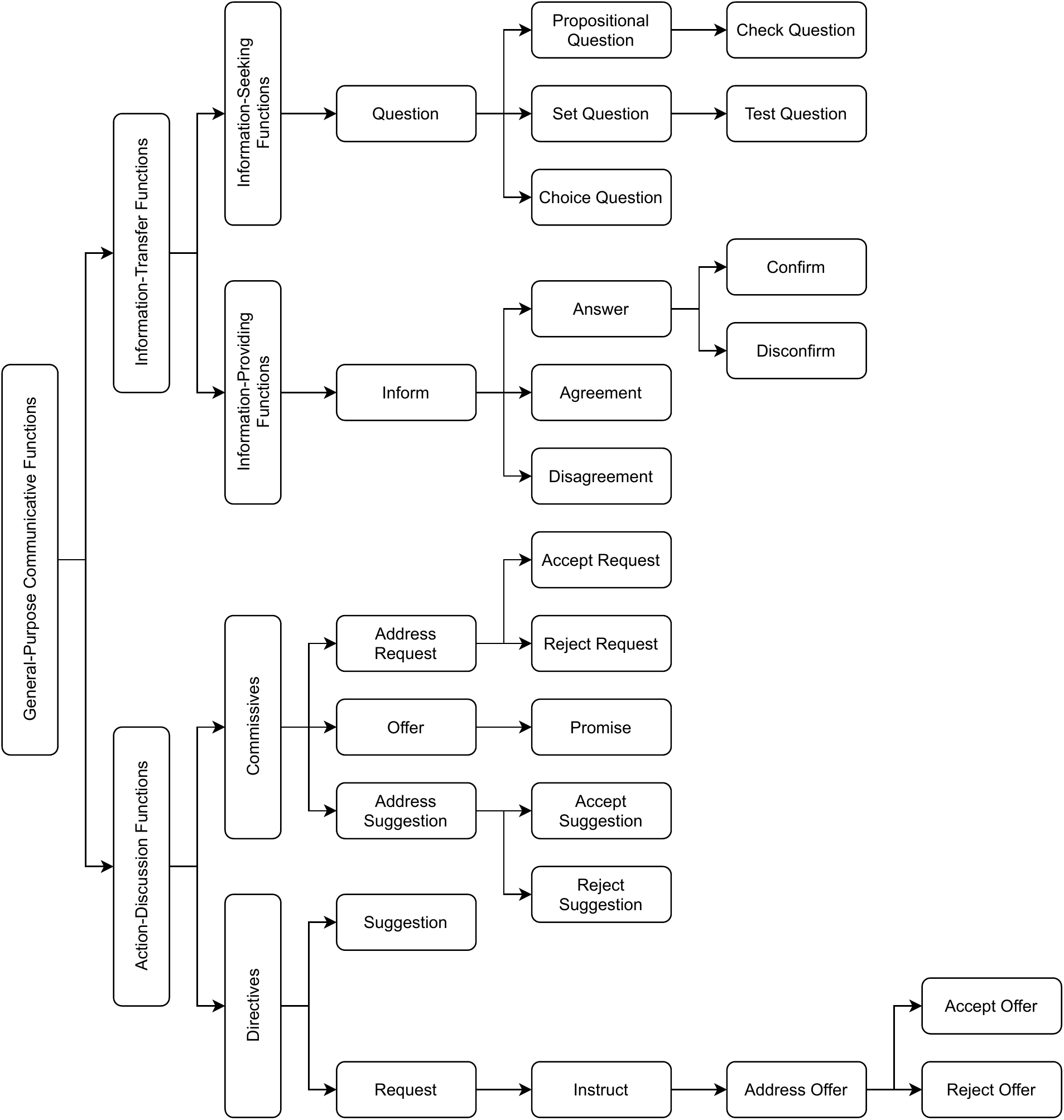}
    \caption{Hierarchy of general-purpose communicative functions defined by the ISO 24617-2 standard for dialog act annotation.}
    \label{fig:generalpurpose}
\end{figure}

The standard defines nine semantic dimensions {--} \textit{Task}, \textit{Auto-Feedback}, \textit{Allo-Feedback}, \textit{Turn Management}, \textit{Time Management}, \textit{Discourse Structuring}, \textit{Own Communication Management}, \textit{Partner Communication Management}, and \textit{Social Obligations Management} {--} in which different communicative functions may occur. These communicative functions are the standard equivalent of the dialog act labels used to annotate dialogs before the introduction of the standard. They are divided into general-purpose and dimension-specific functions. The former can occur in any semantic dimension and are organized hierarchically as shown in Figure~\ref{fig:generalpurpose}. The latter can only occur in the corresponding dimension. The latter can only occur in the corresponding dimension and are distributed as shown in Table~\ref{tab:dimensionspecific}. Of the nine semantic dimensions, only the \textit{Task} dimension does not have specific functions. This means that only general-purpose communicative functions occur in that dimension. Furthermore, with the exception of the functions specific to the \textit{Social Obligations Management} dimension, which can be split into their initial and return counterparts, dimension-specific functions are all at the same level.

\begin{table}
    \centering
    \begin{tabular}{l l}
        \toprule
        \textbf{Semantic Dimension} & \textbf{Communicative Functions} \tabularnewline
        \midrule
        \multirow{2}{*}{Auto-Feedback}
        & Auto Positive \tabularnewline
        & Auto Negative \tabularnewline
        \midrule
        \multirow{3}{*}{Allo-Feedback}
        & Allo Positive \tabularnewline
        & Allo Negative \tabularnewline
        & Feedback Elicitation \tabularnewline
        \midrule
        \multirow{3}{*}{Own Communication Management}
        & Retraction \tabularnewline
        & Self Correction \tabularnewline
        & Self Error \tabularnewline
        \midrule
        \multirow{2}{*}{Partner Communication Management}
        & Correct Misspeaking \tabularnewline
        & Completion \tabularnewline
        \midrule
        \multirow{6}{*}{Turn Management}
        & Turn Accept  \tabularnewline
        & Turn Assign \tabularnewline
        & Turn Grab \tabularnewline
        & Turn Keep \tabularnewline
        & Turn Release \tabularnewline
        & Turn Take \tabularnewline
        \midrule
        \multirow{2}{*}{Time Management}
        & Stalling \tabularnewline
        & Pausing \tabularnewline
        \midrule
        \multirow{2}{*}{Discourse Structuring}
        & Interaction Structuring \tabularnewline
        & Opening \tabularnewline
        \midrule
        \multirow{5}{*}{Social Obligations Management}
        & Greeting \tabularnewline
        & Self Introduction \tabularnewline
        & Apology \tabularnewline
        & Thanking \tabularnewline
        & Goodbye \tabularnewline
        \bottomrule
    \end{tabular}
    \caption{Dimension-specific communicative functions defined by the ISO 24617-2 standard for dialog act annotation.}
    \label{tab:dimensionspecific}
\end{table}

Annotating all of the aspects defined by the standard is an exhaustive process and, consequently, the amount of available data is still reduced and, in many cases, not all of the aspects are considered~\cite<e.g.>{Petukhova2014,Bunt2016,Bunt2019,Anikina2019}. To the best of our knowledge, the DialogBank~\cite{Bunt2016,Bunt2019} is the only publicly available source of dialogs fully annotated according to the standard. It features (re)-annotated dialogs from several corpora, but, currently, there are only 15 dialogs in English and 9 in Dutch, which amount to less than 3,000 segments.

\subsection{Dialog Act Recognition}
\label{sec:related:recognition}

Given a turn, utterance, or functional segment in a dialog, to which we will refer generically as segment in the remainder of the article, automatic dialog act recognition aims at identifying the intention behind that segment. It is a task that has been widely explored over the years, using both classical machine learning and deep learning approaches. In both cases, the approaches differ mainly on how the representation of a segment is generated from the representations of its tokens and how they are able to weigh context information in the decision process. The article by \citeA{Kral2010} provides a comprehensive overview on classical machine learning approaches on the task, except for the more recent \ac{SVM}-based approaches~\cite{Gamback2011,Ribeiro2015}. 

Regarding deep learning approaches, both \acp{RNN}~\cite<e.g>{Lee2016,Ji2016,Khanpour2016,Tran2017a} and \acp{CNN}~\cite<e.g.>{Kalchbrenner2013,Lee2016,Liu2017} have been used to generate segment representations by combining the embedding representations of their words. While the first focus on capturing information from relevant sequences of tokens, the latter focus on the context surrounding each token and, thus, on capturing relevant token patterns independently of where they occur in the segment. We have compared different \ac{RNN}- and \ac{CNN}-based segment representation approaches and achieved higher performance on the task using a set of parallel \acp{CNN} with different window sizes, while also consuming less resources than when using \acp{RNN}~\cite{Ribeiro2019b}. Still, most of the more recent studies on the task rely on bidirectional \acp{RNN} for segment representation~\cite<e.g.>{Kumar2018,Chen2018,Li2019,Raheja2019}.

Regarding the representation of the segment's tokens, most approaches on dialog act recognition using deep learning have relied on pre-trained word embedding representations generated by Word2Vec~\cite{Mikolov2013} or GloVe~\cite{Pennington2014}. However, similarly to what happens on most \ac{NLP} tasks, using contextualized word representations, especially those generated by BERT~\cite{Devlin2019}, leads to higher performance~\cite{Ribeiro2019b}. Additionally, there are studies which also rely on character-level information \cite<e.g.>{Bothe2018,Chen2018,Li2019,Raheja2019}. In our studies on that matter, we observed that there are complementary cues for intention at a sub-word level which cannot be captured when relying solely on word-level tokenization~\cite{Ribeiro2018a,Ribeiro2019a}. Finally, functional-level tokenization, mainly in the form of \ac{POS} tags, has also been explored for the task~\cite{Chen2018,Ribeiro2019b}. However, it seems to be less relevant.

In terms of context information, previous studies have observed performance improvements when using information regarding turn-taking by the speakers~\cite{Liu2017,Ribeiro2019b} and the topics covered by the dialog~\cite{Li2019}. Still, the most important source of context information is the surrounding segments. Studies dedicated to that matter have shown that the influence of preceding segments decreases with the distance and that their dialog act classification is more informative than their words, even when the classifications are obtained automatically~\cite{Ribeiro2015,Liu2017}. Furthermore, we have shown that sequentiality information and long distance dependencies between the preceding segments can be captured by using a \ac{RNN} to generate a summary of their classifications~\cite{Ribeiro2019b}. 

In fact, considering the dependencies between the multiple segments in a dialog, several studies attempted to predict the sequence of dialog acts in a complete dialog by approaching the task as a sequence labeling problem~\cite<e.g.>{Bothe2018,Kumar2018}. These cases rely on a hierarchical approach in which the representations of the segments are provided to a conversation-level \ac{RNN} that models the whole dialog. Furthermore, the performance can be improved by including attention mechanisms that identify the information in the surrounding segments that is most relevant for predicting the dialog acts~\cite<e.g.>{Tran2017a,Chen2018,Li2019,Raheja2019}. \citeA{Tran2017c} also observed that propagating uncertainty information concerning the previous predictions can lead to the prediction of better dialog act sequences. Finally, most of the studies that approach the task as a sequence labeling problem also rely on \acp{CRF}~\cite<e.g.>{Kumar2018,Chen2018,Li2019,Raheja2019} or generative models~\cite{Tran2017b} as the final layer to predict the sequence of dialog acts. This way, the prediction of the dialog act for a segment is further conditioned to the previous predictions. Overall, this kind of approach achieves the highest performance on the task. However, the conversation-level \ac{RNN} is typically bidirectional. This means that the prediction of the dialog act for a segment relies not only on the information from previous segments, but also from future ones, which are not available to a dialog system.

Additional studies on dialog act recognition explore alternative approaches or focus on specific applications. For instance, \citeA{Wan2018} approached the task as a \ac{QA} problem and employed adversarial training. On the other hand, \citeA{Ravi2018} focused on developing dialog act recognition models that can be used in mobile applications.

Regarding the recognition of the communicative functions defined by the ISO 24617-2 standard, to the best of our knowledge, besides us, only \citeA{Anikina2019} have addressed that problem. More specifically, they have explored the recognition of a compressed set of eight communicative functions on the TRADR corpus~\cite{KruijffKorbayova2015}. This compressed set merges functions in the \textit{Task}, \textit{Turn Management}, and \textit{Feedback} dimensions and does not consider the hierarchical nature of general-purpose functions. Thus, the task was approached as a flat classification problem. The authors compared the performance of several \ac{DNN} architectures and uncontextualized embedding approaches and observed the highest performance when the representation of the segment was generated by passing GloVe embeddings through a \ac{LSTM}. However, the use of parallel \acp{CNN} was not explored and only large window sizes were considered. Furthermore, similarly to what was observed in previous studies on dialog act recognition, using context information regarding the dialog history led to improved performance. However, in this case, it was summarized as the average of the embedding representations of all the words in the dialog history.

Given the reduced amount of dialogs in the DialogBank, we have mapped the dialog act labels of the LEGO corpus~\cite{Schmitt2012} into the communicative functions defined by the ISO 24617-2 standard. To assess the utility of these partially annotated dialogs, we have performed preliminary experiments on the automatic recognition of general-purpose communicative functions in the DialogBank~\cite{Ribeiro2020}. In those experiments, we flattened the hierarchy and addressed the task as a flat dialog act recognition problem using the same approach we used to predict the original dialog act labels of LEGO corpus~\cite{Ribeiro2019a}. Then, we observed performance improvements when the LEGO dialogs were included in the training phase, in comparison to when the classifier was trained solely on DialogBank dialogs.

%
%

\section{General-Purpose Communicative Function Recognition}
\label{sec:approach}

The main difference between general-purpose communicative function recognition and traditional dialog act recognition is that the former poses a hierarchical classification problem, with paths that may not end on a leaf. However, both are intention recognition problems at their core. Thus, we approach the problem by adapting existing dialog act recognition approaches to deal with hierarchical problems. This way, we build on the ability of those approaches to capture generic information regarding intention. Furthermore, this allows us to explore the use of existing pre-trained models in transfer learning processes, in an attempt to minimize the impact of the scarcity of annotated data.

\begin{figure}
    \centering
    \includegraphics[width=0.9\textwidth]{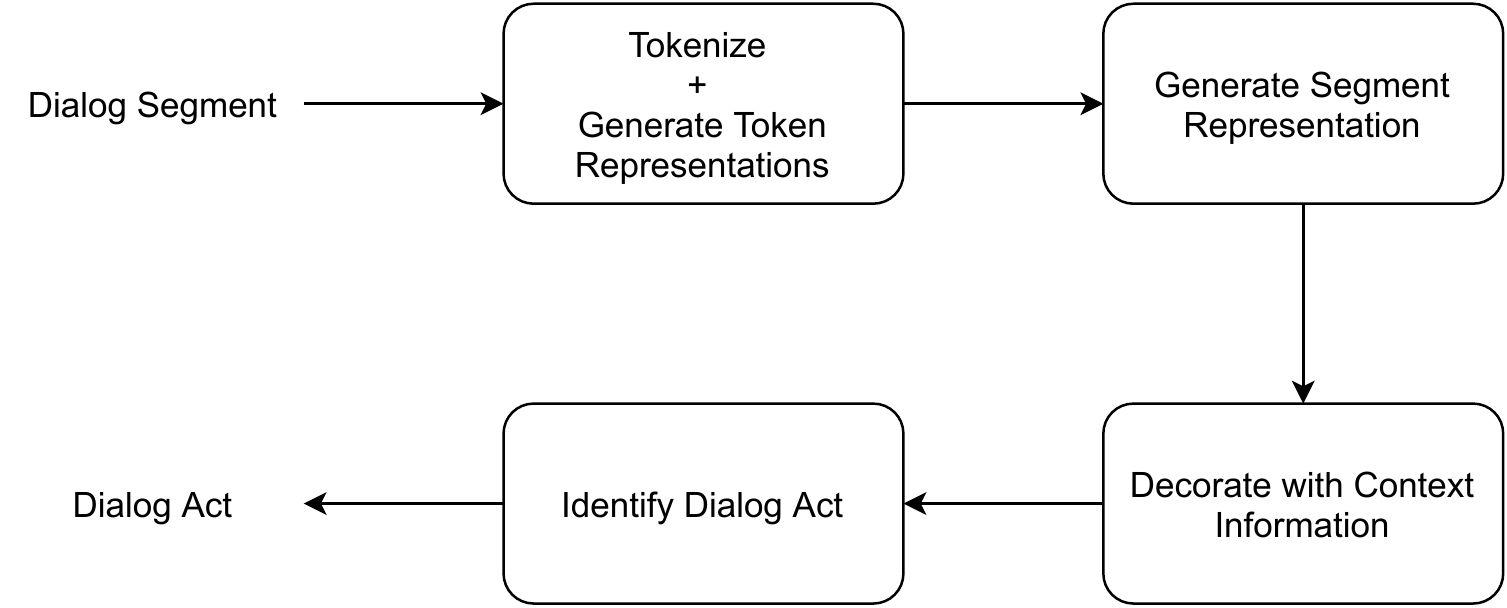}
    \caption{The generic dialog act recognition process.}
    \label{fig:approach:generic}
\end{figure}

Although the studies on dialog act recognition covered in Section~\ref{sec:related:recognition} explored different aspects that are relevant for the task, most approaches can be summarized as the generic four-step process shown in Figure~\ref{fig:approach:generic}. Given a segment in a dialog, the first step towards the identification of the main dialog act that it communicates is to split it into its constituent tokens and generate adequate representations for each of them. As discussed in Section~\ref{sec:related:recognition}, the tokenization is typically performed at the word-level. However, other tokenizations, such as at the character or functional levels, can be used to provide complementary information. In such cases, all the different tokenizations of the segment are considered in the subsequent steps of the process. The next step is to generate a representation of the segment by combining the representations of its tokens. Ideally, this representation should focus on capturing the characteristics of the segment that are relevant to identify the intention that it transmits. Then, the representation is decorated with context information regarding the dialog history and speaker information. This context information can be provided in the form of external features or by the recognition process itself in a recurrent manner. Finally, the information provided by the decorated segment representation is used to identify the dialog act communicated by the segment. 
 
Since the segment representation decorated with context information focuses on providing information that allows the identification of the intention behind the segment, all the steps towards its generation are relevant for the recognition of both traditional dialog acts and general-purpose communicative functions. Consequently, the adaptation of existing dialog act recognition approaches to the recognition of general-purpose communicative functions refers to how that decorated segment representation can be specialized to allow the identification of the hierarchically structured functions. A possible approach, which does not require modifications to the architecture of existing dialog act recognition approaches, is to simply flatten the hierarchy. This way, the recognition of general-purpose communicative functions can be approached as a flat single-label classification problem. However, this means that the relations between each communicative function and its ancestors and descendants are not considered. Thus, we use it as a baseline for assessing the ability of our approach to capture information regarding hierarchical dependencies and use it to improve the ability to recognize general-purpose communicative functions.

\begin{figure}
    \centering
    \includegraphics[width=\textwidth]{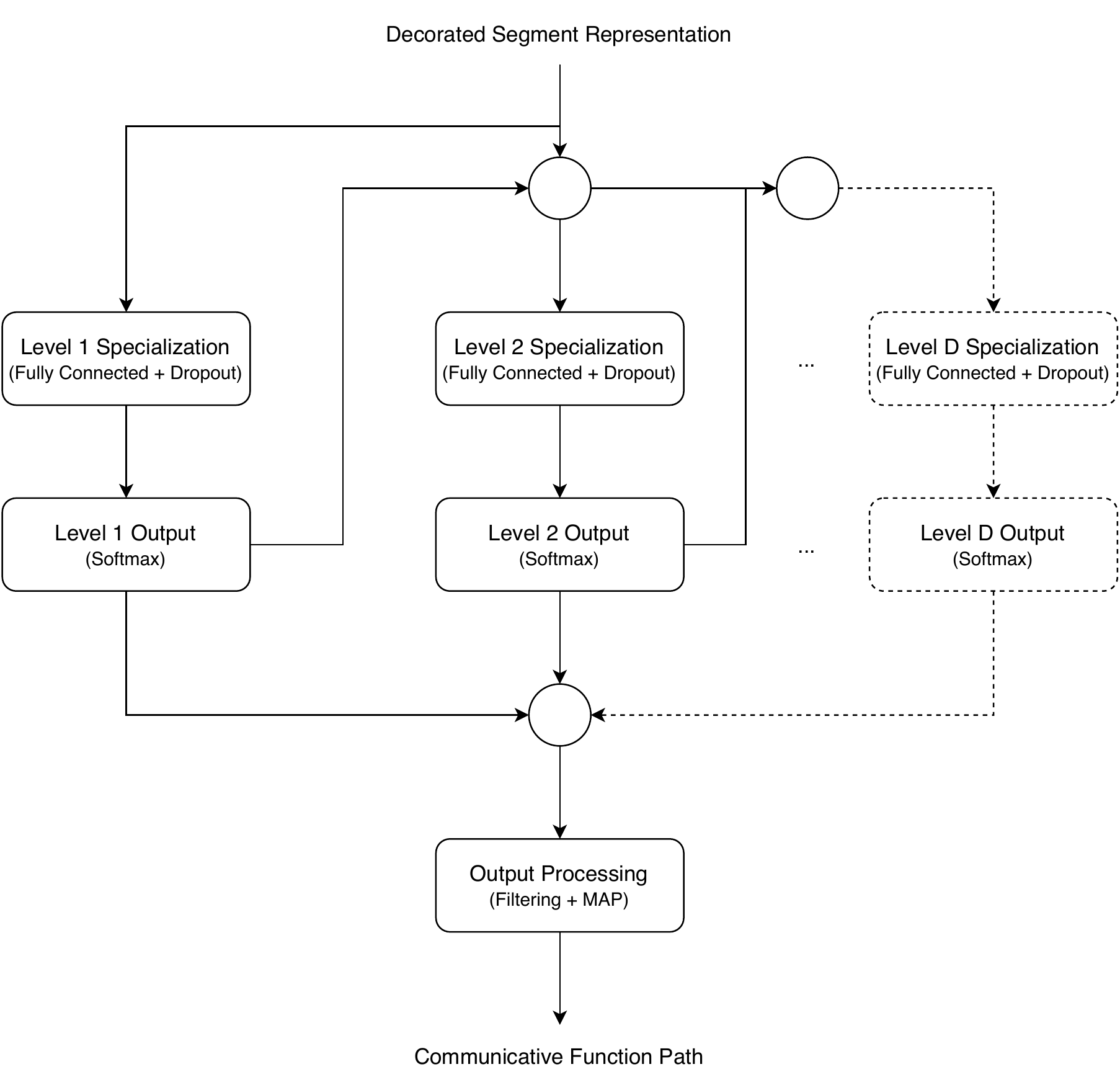}
    \caption{Our adaptation of a generic dialog act recognition approach to deal with the hierarchical problem posed by the ISO 24617-2 general-purpose communicative functions. The input is a segment representation decorated with context information. The circles represent concatenation operations.}
    \label{fig:approach:hierarchical}
\end{figure}

The top performing dialog act recognition approaches are based on deep neural networks. In this context, when dealing with the multi-class single-label classification problems posed by most dialog act annotations, the output layer applies the \textit{softmax} activation function to obtain a probability distribution of the classes. The dialog act of a given segment is then predicted by selecting the class with highest probability. As shown in Figure~\ref{fig:approach:hierarchical}, in order to consider the hierarchical structure of the general-purpose communicative functions defined by the ISO 24617-2 standard, we propose to use an output layer per level of the hierarchy instead of using a single output layer. This way, each output layer focuses on distinguishing between communicative functions at the corresponding level without having to deal with the ambiguity caused by functions that are ancestors or descendants of each other.

Additionally, we introduce a specialization layer per level, which is a fully connected layer that, as the name suggests, specializes the decorated segment representation by capturing the information that is most relevant for distinguishing the communicative functions at that level of the hierarchy. Furthermore, these layers are used to reduce the probability of overfitting by applying dropout~\cite{Srivastava2014} during the training phase. The use of specialization layers has already been proved important in our studies on automatic dialog act recognition~\cite{Ribeiro2018b,Ribeiro2019b}, including those on the DIHANA corpus~\cite{Benedi2006}, which is annotated for dialog acts at three different levels.

The final modification to existing network architectures refers to the use of cascading outputs, that is, the probability distribution predicted by the network at a given level is appended to the decorated segment representation before it is passed to the specialization layer of the next level. This way, the network can capture information regarding the hierarchical dependencies between communicative functions at different levels.

The general-purpose communicative functions defined by the ISO 24617-2 standard follow a strict hierarchy, the classification of a segment does not necessarily end on a leaf, and the leaves are not all at the same level. Thus, the network must be able to predict paths with variable length. To approach this problem, we add an additional label, \textit{None}, to each level of the hierarchy, to represent that there is no communicative function attributed to the segment at that level. This way, we are able to simulate paths with fixed length, while introducing minimal impact on the network during the training phase. The drawback is that the \textit{None} label may become the predominant one in the deeper levels, biasing the output layers towards its prediction. These additional labels are also considered when providing context information to the network, in order to have fixed dimensionality.

During the inference phase, the parent-child relations between the general-purpose communicative functions must be considered in order to avoid predicting an invalid path. This means that, when selecting the label at a given level, only the children of the label selected for the level above it can be considered. This restriction can be enforced using an iterative approach that starts by selecting the communicative function with highest probability at the top level and then applies a mask on the predictions of the level below it, in order to discard the communicative functions that are not children of the selected one. This process is then repeated for each level of the hierarchy. However, the performance of this approach is highly impaired when misclassifications occur in the upper levels of the hierarchy. 

To attenuate the impact of misclassifications in the upper levels of the hierarchy, we explore a prediction approach based on \ac{MAP} estimation in which the predicted communicative function for a given segment, $s$, is given by

\begin{equation}
    \text{CommunicativeFunction}(s) = \argmax_{f \in F} \prod_{d=1}^D P(L_d = \text{Path}(f)_d \:\vert\: s),
\end{equation}

\noindent
where $F$ is the set of general-purpose communicative functions, $D$ is the depth of the hierarchy, and $P(L_d=c \:\vert\: s)$ is given by the softmax output corresponding to label $c$ of the output layer corresponding to the level at depth $d$. That is, instead of iteratively selecting a communicative function at each level, we compute the posterior probability of all valid paths in the hierarchy and select that with highest probability.

Since the classification of a segment does not necessarily end on a leaf and the leaves are not all at the same level, we also rely on the additional label \textit{None} labels during the inference phase. For instance, given a segment $s$, the probability of selecting the path that ends in the \textit{Answer} communicative function is given by

\begin{equation}
    \begin{split}
        P(Answer\:\vert\: s) & = P(L_1=\textit{Information-Transfer Functions} \:\vert\: s) \\ 
                    & \times P(L_2=\textit{Information-Providing Functions} \:\vert\: s) \\
                    & \times P(L_3=Inform \:\vert\: s) \\
                    & \times P(L_4=Answer \:\vert\: s) \\
                    & \times P(L_5=None \:\vert\: s) \\
                    & \times P(L_6=None \:\vert\: s).
    \end{split}
\end{equation}

%
%

\section{Experimental Setup}
\label{sec:setup}

This section describes our experimental setup, including the datasets, the evaluation methodology, and the complete network architecture used in our experiments, including implementation details that allow future reproduction of those experiments.

\subsection{Datasets}
\label{ssec:datasets}

As discussed in Section~\ref{sec:related:standard}, to the best of our knowledge, the DialogBank~\cite{Bunt2016,Bunt2019} is the only publicly available source of dialogs annotated fully according to the ISO 24617-2 standard guidelines. Thus, in our experiments, we use it as gold standard for evaluating the performance of the different approaches. However, given the scarcity of annotated dialogs, we also rely on the LEGO-ISO dataset~\cite{Ribeiro2020}, which features dialogs that were partially annotated with the communicative functions defined by the ISO 24617-2 standard through label mapping processes. Finally, we also rely on Switchboard Dialog Act Corpus~\cite{Jurafsky1997}, which is the most explored in dialog act recognition studies, to train models for transfer learning purposes. These datasets are described in further detail below.

\subsubsection{The DialogBank}

The DialogBank~\cite{Bunt2016,Bunt2019} aims at collecting and providing dialogs annotated fully according to the ISO 24617-2 standard guidelines. At the time of this study, it features (re-)annotated dialogs from four English corpora and four Dutch corpora. There is a total of 15 annotated dialogs in English and 9 in Dutch. To avoid the issues regarding multilinguality, we focus on the English dialogs in this study. Three of those dialogs are originally from MapTask~\cite{Anderson1991}, four are from Switchboard~\cite{Godfrey1992}, three are from TRAINS~\cite{Allen1991}, and five are from DBOX~\cite{Petukhova2014}. In total, the dialogs contain 2,360 annotated segments, out of which 1,118 have general-purpose communicative functions in the \textit{Task} dimension.

\begin{table}
    \centering
    \begin{tabular}{l r r r r r}
        \toprule
        \textbf{Function}      & \textbf{MapTask} & \textbf{Switchboard} & \textbf{TRAINS} & \textbf{DBOX} & \textbf{Total} \tabularnewline
        \midrule
        Inform                 &               56 &                  338 &              44 &            37 &            475 \tabularnewline
        Instruct               &              143 &                    0 &               1 &            11 &            155 \tabularnewline
        Answer                 &               35 &                   30 &              16 &            31 &            112 \tabularnewline
        Propositional Question &               26 &                   11 &               2 &            25 &             64 \tabularnewline
        Set Question           &                7 &                   12 &              13 &            28 &             60 \tabularnewline
        Accept Request         &               52 &                    0 &               1 &             1 &             54 \tabularnewline
        Agreement              &                8 &                   42 &               2 &             1 &             53 \tabularnewline
        Check Question         &               28 &                    9 &               7 &             6 &             50 \tabularnewline
        Confirm                &               11 &                    9 &               6 &            14 &             40 \tabularnewline
        Suggest                &                3 &                    3 &               2 &             5 &             13 \tabularnewline
        Disconfirm             &                1 &                    1 &               0 &            10 &             12 \tabularnewline
        Request                &                2 &                    2 &               1 &             4 &              9 \tabularnewline
        Choice Question        &                3 &                    0 &               1 &             4 &              8 \tabularnewline
        Correction             &                2 &                    0 &               0 &             1 &              3 \tabularnewline
        Address Request        &                3 &                    0 &               0 &             0 &              3 \tabularnewline
        Offer                  &                0 &                    0 &               0 &             2 &              2 \tabularnewline
        Decline Offer          &                0 &                    0 &               0 &             1 &              1 \tabularnewline
        Disagreement           &                1 &                    0 &               0 &             0 &              1 \tabularnewline
        Accept Offer           &                0 &                    0 &               0 &             1 &              1 \tabularnewline
        Accept Suggest         &                0 &                    0 &               0 &             1 &              1 \tabularnewline
        Promise                &                0 &                    0 &               0 &             1 &              1 \tabularnewline
        \midrule
        \textbf{General-Purpose CFs}         &              381 &                  457 &              96 &           184 &           1,118 \tabularnewline

        \textbf{None}          &              281 &                  555 &             140 &           266 &           1,242 \tabularnewline
        \midrule
        \textbf{Total}         &              662 &                1,012 &             236 &           450 &             2,360 \tabularnewline 
        \bottomrule
    \end{tabular}
    \caption{Distribution of the general-purpose communicative functions defined by the ISO 24617-2 standard in the DialogBank.}
    \label{tab:setup:datasets:dialogbank}
\end{table}

The general-purpose communicative functions are distributed in the DialogBank according to Table~\ref{tab:setup:datasets:dialogbank}. We can see that, overall, the distribution is highly unbalanced, with the most common, \textit{Inform}, covering 42\% of the segments that have a general-purpose communicative function, while 10 of the functions that occur in the DialogBank occur in less than 10 segments. The predominance of the \textit{Inform} communicative function becomes even more apparent if we consider the paths in the hierarchy. \textit{Answer}, \textit{Agreement}, \textit{Confirm}, \textit{Disconfirm}, and \textit{Disagreement} are descendants of \textit{Inform}. This means that of the segments that have a general-purpose communicative function, 62\% have the \textit{Inform} function.

Another important aspect revealed in Table~\ref{tab:setup:datasets:dialogbank} is that the distribution of general-purpose communicative functions is also highly unbalanced across the dialogs of the different corpora that are included in the DialogBank, even in terms of the most common communicative functions. For instance, not considering paths, 92\% of the segments with the \textit{Instruct} and 96\% with the \textit{Accept Request} communicative functions belong to MapTask dialogs. On the other hand, 71\% of the segments with the \textit{Inform} communicative function belong to Switchboard dialogs. This reveals the heterogeneity of the DialogBank which is representative of the different kinds of dialog that occur in human-human and human-machine interaction. 

Overall, although it includes dialogs from multiple corpora, the amount of data provided by the DialogBank is not enough for drawing solid conclusions from the results of \ac{DNN}-based approaches trained solely on it, especially considering the hierarchical nature of the general-purpose communicative functions that we intend to recognize automatically. However, since these dialogs are the closest we have to a gold standard annotation, the evaluation of our approaches on general-purpose communicative function recognition must be based on the performance on the DialogBank.

\subsubsection{LEGO-ISO}

LEGO-ISO~\cite{Ribeiro2020} consists of 347 dialogs from the Let’s Go Bus Information System~\cite{Raux2006}, containing 14,186 utterances annotated with the communicative functions defined by the ISO 24617-2 standard. Each dialog features the system and a human user. Since the 9,803 system utterances are generated through slot filling of fixed templates, they have no errors and contain casing and punctuation information. In contrast, the transcriptions of the 5,103 user utterances were obtained using an \ac{ASR} system and, consequently, are subject to recognition errors and contain no casing nor punctuation information. However, a concrete value for the transcription \ac{WER} was not revealed. 

The annotation with the standard's communicative functions was obtained through the mapping of the original dialog act annotations of the LEGO corpus~\cite{Schmitt2012}. The mapping was based solely on the original labels and the transcriptions of the turns. This means that the annotation is performed on turns rather than on functional segments and that it does not cover every semantic dimension. Table~\ref{tab:setup:datasets:lego} shows the distribution of general-purpose communicative functions in the corpus after the mapping. We can see that, although the number of segments is larger, the set of communicative functions covered by the corpus is only a subset of that covered by the DialogBank. This is partially due to the label mapping process, which did not consider the specificities of certain segments, but, most importantly, it is due to the characteristics of the dialogs, which are highly focused on the task. Thus, system segments typically have a communicative function that is a descendant of \textit{Question}, so that it can obtain all the information required to fulfill the task. On the other hand, user segments typically have a communicative function that is a descendant of \textit{Inform}, since they aim at providing that information to the system. Furthermore, the most common communicative function is \textit{Check Question} because the system tries to confirm that it understood every piece of information provided by the user correctly. Also given the focus on the task, only 17\% of the segments do not have a general-purpose communicative function, which contrasts with the 53\% of the DialogBank.

\begin{table}
    \centering
    \begin{tabular}{l r r r}
        \toprule
        \textbf{Function}            & \textbf{System Segments} & \textbf{User Segments} & \textbf{Total} \tabularnewline
        \midrule
        Check Question               &                    2,256 &                      1 &          2,257 \tabularnewline
        Set Question                 &                    1,987 &                    210 &          2,197 \tabularnewline
        Instruct                     &                    1,812 &                    106 &          1,918 \tabularnewline
        Answer                       &                        0 &                  1,462 &          1,462 \tabularnewline
        Inform                       &                      656 &                    600 &          1,256 \tabularnewline
        Confirm                      &                        0 &                  1,162 &          1,162 \tabularnewline
        Disconfirm                   &                        0 &                  1,105 &          1,105 \tabularnewline
        Promise                      &                      277 &                      0 &            277 \tabularnewline
        Request                      &                       54 &                     85 &            139 \tabularnewline
        Suggest                      &                       40 &                      0 &             40 \tabularnewline
        \midrule         
        \textbf{General-Purpose CFs} &                    7,082 &                  4,731 &         11,813 \tabularnewline
        \textbf{None}                &                    2,001 &                    372 &          2,373 \tabularnewline
        \midrule         
        \textbf{Total}               &                    9,083 &                  5,103 &         14,186 \tabularnewline
        \bottomrule
    \end{tabular}
    \caption{Distribution of the general-purpose communicative functions defined by the ISO 24617-2 standard in the LEGO-ISO corpus.}
    \label{tab:setup:datasets:lego}
\end{table}

Since its communicative-function annotations were obtained through label mapping processes, this dataset cannot be used as a gold standard. Still, it is 20 times larger than the DialogBank in number of English dialogs and 6 times larger in number of annotated segments. Thus, it provides a significant amount of data that, according to the results of our preliminary studies~\cite{Ribeiro2020}, can be used during the training phase to improve the performance on general-purpose communicative function recognition. However, given its size in comparison to the DialogBank, classifiers may overfit to its characteristics.

\subsubsection{Switchboard Dialog Act Corpus}

The Switchboard Dialog Act Corpus~\cite{Jurafsky1997} is an annotated subset of the Switchboard~\cite{Godfrey1992} corpus. It is the largest and most explored corpus annotated with dialog act information, consisting of 1,155 manually transcribed conversations, containing 223,606 segments. The conversations are between pairs of humans and cover multiple domains. The corpus is annotated for dialog acts using the domain-independent SWBD-DAMSL tag set, which features over 200 unique labels. However, most studies use a reduced set of 42 to 44 labels to obtain a higher inter-annotator agreement and higher example frequencies per class. Table~\ref{tab:setup:datasets:swda} shows this set of labels and its distribution in the corpus. The total number of segments labeled with a dialog act is lower than the previously referred 223,606 since some segments are considered continuations of the previous one by the same speaker and aggregated to it.

\begin{table}[ht]
    \centering
    \begin{tabular}{l r l r}
        \toprule
        \textbf{Label}              & \textbf{Count} & \textbf{Label}           & \textbf{Count} \tabularnewline
        \midrule
        Statement-Non-Opinion       &         72,824 & Collaborative Completion &            699 \tabularnewline
        Acknowledgement             &         37,096 & Repeat-Phrase            &            660 \tabularnewline
        Statement-Opinion           &         25,197 & Open-Question            &            632 \tabularnewline
        Agreement                   &         10,820 & Rhetorical-Question      &            557 \tabularnewline
        Abandoned                   &         10,569 & Hold                     &            540 \tabularnewline
        Appreciation                &          4,663 & Reject                   &            338 \tabularnewline
        Yes-No-Question             &          4,624 & Negative Non-No Answer   &            292 \tabularnewline
        Non-Verbal                  &          3,548 & Non-understanding        &            288 \tabularnewline
        Yes Answer                  &          2,934 & Other Answer             &            279 \tabularnewline
        Conventional Closing        &          2,486 & Conventional Opening     &            220 \tabularnewline
        Uninterpretable             &          2,158 & Or-Clause                &            207 \tabularnewline
        Wh-Question                 &          1,911 & Dispreferred Answers     &            205 \tabularnewline
        No Answer                   &          1,340 & 3rd-Party-Talk           &            115 \tabularnewline
        Response Acknowledgement    &          1,277 & Offers / Options         &            109 \tabularnewline
        Hedge                       &          1,182 & Self-talk                &            102 \tabularnewline
        Declarative Yes-No-Question &          1,174 & Downplayer               &            100 \tabularnewline
        Other                       &          1,074 & Maybe                    &             98 \tabularnewline
        Backchannel-Question        &          1,019 & Tag-Question             &             93 \tabularnewline
        Quotation                   &            934 & Declarative Wh-Question  &             80 \tabularnewline
        Summarization               &            919 & Apology                  &             76 \tabularnewline
        Affirmative Non-Yes Answer  &            836 & Thanking                 &             67 \tabularnewline
        Action Directive            &            719 &                          &                \tabularnewline
        \midrule
        \textbf{Total}              &                &                          &        195,061 \tabularnewline
        \bottomrule
    \end{tabular}
    \caption{Dialog act distribution in the Switchboard Dialog Act Corpus.}
    \label{tab:setup:datasets:swda}
\end{table}

Although the corpus is not annotated according to the ISO 24617-2 standard, its dialog act annotations and the communicative functions defined by the standard reveal similar intentions. For instance, regarding general-purpose communicative functions, labels such as \textit{Yes-No-Question}, \textit{Yes Answer}, and \textit{No Answer}, can be directly mapped into the \textit{Propositional Question}, \textit{Confirm}, and \textit{Disconfirm} communicative functions, respectively. Mappings of this kind are possible for several other labels, not only to general-purpose communicative functions, but also to dimension-specific functions. \citeA{Fang2012} provide further insight into the possible mapping between the dialog act labels of the Switchboard Dialog Act Corpus and the communicative functions defined by the ISO 24617-2 standard.

Given the size of the corpus and the similarity of the intentions revealed by its dialog act label set, we use it in our experiments to train a flat dialog act recognition model, so that its weights can be used in transfer learning processes. This way, the probability of overfitting the representations to the characteristics of the training dialogs is reduced, which may improve the generalization ability of the classifiers.
\subsection{Evaluation Methodology}
\label{sec:setup:evaluation}

In this study, we focus on the recognition of ISO 24617-2 general-purpose communicative functions in the DialogBank. Below, we describe the evaluation scenarios, the evaluation approach, and the metrics. 

\subsubsection{Scenarios}

Although general-purpose communicative functions may occur in any of the semantic dimensions defined by the ISO 24617-2 standard, we focus on the \textit{Task} dimension, since the number of occurrences of general-purpose communicative functions in the remaining dimensions is not representative in the DialogBank. In this context, we defined two evaluation scenarios. The first focuses on the recognition of the different general-purpose functions in the segments that have communicative functions in the \textit{Task} dimension. Thus, the remaining segments are discarded. On the other hand, the second scenario also considers the identification of segments which have communicative functions in the \textit{Task} dimension. Thus, all segments are considered and a new label, \textit{None}, is given to those which do not have communicative functions in that dimension.

\subsubsection{Cross-Validation}

Given the reduced amount of dialogs in the DialogBank, it is not feasible to split it into partitions for training, development, and testing. Thus, we evaluate performance using two cross-validation approaches. The first is leave-one-dialog-out cross-validation, that is, the predictions for the segments in each dialog are made by classifiers trained on all the remaining dialogs. We use this as our main evaluation approach because it maximizes the amount of gold standard data available for training.

The second evaluation approach, leave-one-corpus-out cross-validation, takes advantage of the fact the DialogBank features dialogs from multiple corpora. In this case, the predictions for the segments in each dialog do not rely on training information from other dialogs in the same corpus. Thus, to an extent, this approach can be used to assess cross-corpora generalization capabilities.

To keep the evaluation as fair as possible, contrarily to what we did when assessing the ability of the LEGO-ISO dialogs to provide relevant information for training a communicative function recognizer~\cite{Ribeiro2020}, we do not perform any fine tuning to maximize the performance on the left out dialog(s) in each fold. Instead, in each fold, we train an ensemble of classifiers on the corresponding training dialogs. Each of the classifiers in the ensemble is fine-tuned to maximize the performance on one of those training dialogs, while being trained on the remainder and the LEGO-ISO dialogs. This way, we remove the impact of selecting a single dialog for fine-tuning. The predicted classification of each segment in the left out dialog(s) is then given by a weighted majority vote of the classifiers in the ensemble. The weights are given by the estimated probability for the predicted path and ties are broken randomly.

\subsubsection{Metrics}

The most common metric used to evaluate dialog act recognition approaches is accuracy. Its counterpart in the context of hierarchical classification problems is the exact match ratio (MR), which is defined as

\begin{equation}
	\text{MR} = \frac{1}{n}\sum_{i=1}^{n}I(Y_i = Z_i),
\end{equation}

\noindent
where $n$ is the number of evaluation examples, $Y_i$ is the set of labels in the gold standard path of example $i$, $Z_i$ is the set of labels in the path predicted by the classifier for the same example, and $I$ is the indicator function.

In addition to the exact match ratio (MR), we also report results in terms of the hierarchical versions of precision (hP), recall (hR), and F-measure (hF) proposed by \citeA{Kiritchenko2005}, defined as 

\begin{equation}
    \text{hP} = \frac{\sum_{i=1}^{n}|Y_i \cap Z_i|}{\sum_{i=1}^{n}|Z_i|},
\end{equation}

\begin{equation}
    \text{hR} = \frac{\sum_{i=1}^{n}|Y_i \cap Z_i|}{\sum_{i=1}^{n}|Y_i|},
\end{equation}

\begin{equation}
    \text{hF} = \frac{2 * \text{hP} * \text{hR}}{\text{hP} + \text{hR}}.
\end{equation}

\noindent
These hierarchical metrics are relevant, since they consider partial path matches and, thus, capture the difference between predicting a label that shares part of its path with the correct label and one that follows a completely different path.

In some of our experiments we use a set of additional metrics to assess the performance in each level of the hierarchy. Still, these are based on the exact match ratio and on the traditional metrics of precision and recall. To improve readability, we report the values of every metric in percentage form.
\subsection{Network Architecture}
\label{sec:setup:architecture}

As discussed in Section~\ref{sec:approach}, our approach to deal with the hierarchical problem posed by the general-purpose communicative functions defined by the ISO 24617-2 standard can be applied on top of any approach to generate segment representations. In our experiments, we relied on the same approach used in our study that explored the multiple aspects that contribute to dialog act recognition~\cite{Ribeiro2019b}. This way, we know that the segment representation approach is able to capture information regarding intention and we can use the models trained during that study in transfer learning processes.

\begin{figure}
    \centering
    \includegraphics[width=0.9\textwidth]{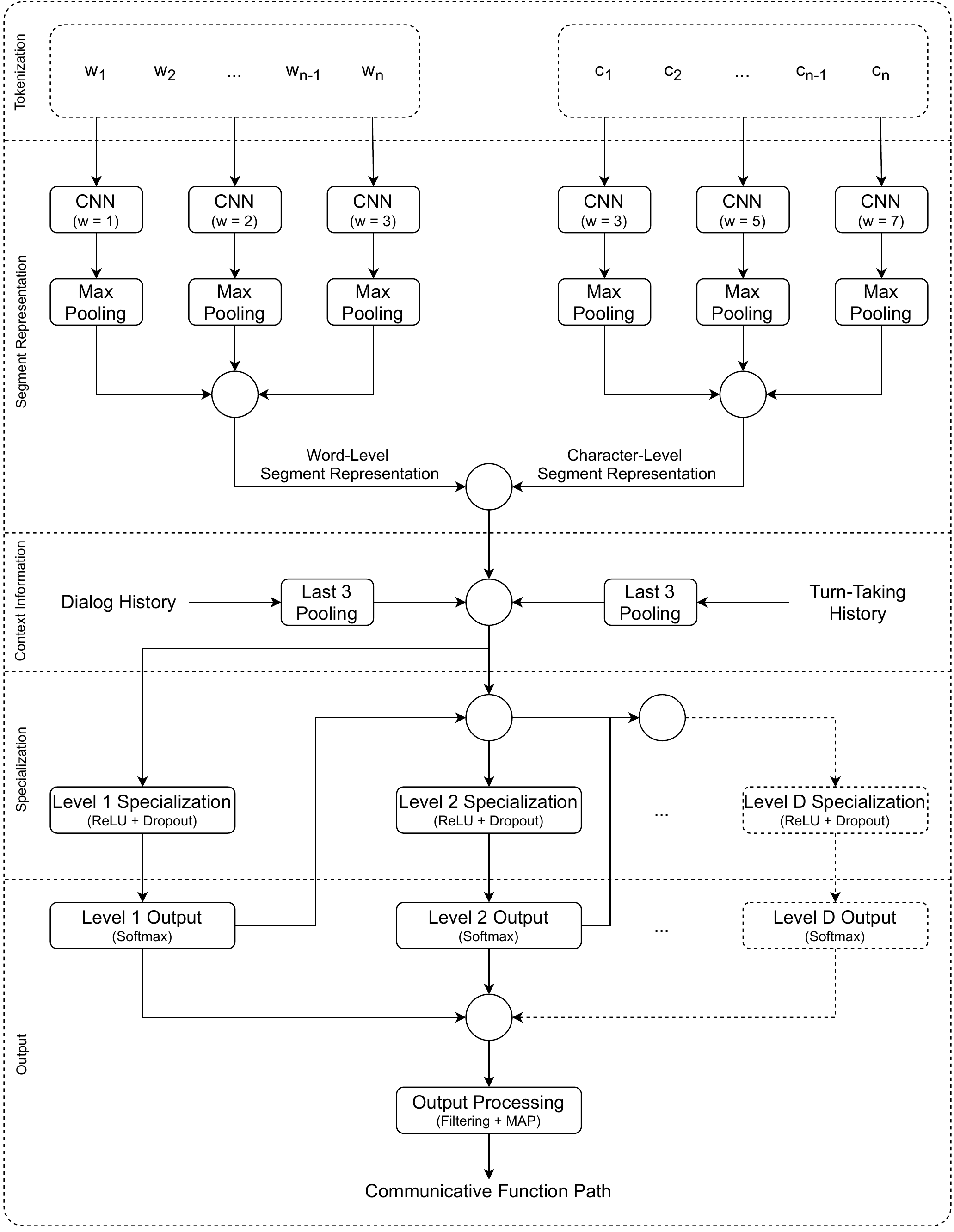}
    \caption{The full architecture of the automatic communicative function recognition approach used in our experiments. $w_n$ and $c_n$ refer to the embedding representation of the $n$-th word and character, respectively. The representations of words are generated by BERT~\cite{Devlin2019} and, thus, are contextualized. $w$ in the \acp{CNN} refers to the width of the context window. $D$ refers to the depth of the hierarchy. The circles represent concatenation operations.}
    \label{fig:setup:architecture}
\end{figure}

Figure~\ref{fig:setup:architecture} shows the complete architecture of the network used in our experiments. Two representations of the segment are generated in parallel, one based on its characters and another on contextualized embedding representations of its words, generated by BERT~\cite{Devlin2019}. In both cases, the representation of the segment is generated by concatenating the outputs of three parallel \acp{CNN} with different window sizes followed by a max-pooling operation. At the character level, we use windows of size three, five, and seven, in order to focus on affixes, lemmas, and inter-word relations. At the word level, we use windows of size one, two, and three, in order to focus on independent words and short word patterns. The two representations are then concatenated and decorated with context information.

Regarding context information, considering that the number of dialogs in the DialogBank is small, we do not rely on a summary of the whole dialog history as in the original approach, because it is prone to overfitting. Instead, we use a flattened sequence of classifications and turn-taking information of the three preceding segments, which have been proved the most important in previous studies~\cite{Ribeiro2015,Liu2017}. The classification of each preceding segment is represented as a concatenation of the one-hot representations of the communicative functions at each level of the hierarchy. Turn-taking information is provided as flags stating whether the speaker changed.

The specialization and output layers are as described in Section~\ref{sec:approach}. That is, there are dedicated specialization and output layers per each level in the hierarchy, the specialization layer of each level also considers the output at the upper levels, and each output layer considers an additional class representing the lack of communicative function at that level. 

In order to take advantage of the intention information captured by a dialog act recognition model trained on a large corpus, we apply a transfer learning process. More specifically, we preset the weights of the segment representation layers of our hierarchical model using the corresponding weights of the flat dialog act recognition model. Consequently, only the specialization and output layers are trained on the dialogs annotated according to the ISO 24617-2 standard. This way, the segment representations provide generic information regarding intention that is then specialized for the distinction among general-purpose communicative functions at each level.
\subsection{Implementation Details}
\label{sec:setup:implementation}

To implement our classifiers, we used Keras~\cite{Chollet2015} with the TensorFlow~\cite{Abadi2015} backend. To update the weights during the training phase, we used the Adam optimizer~\cite{Kingma2015} with the default parameterization and mini-batches with size 512. To decide when to stop training, we used early stopping with 10 epochs of patience. That is, the training phase of each classifier stopped after ten epochs without improvement on the validation set composed of the corresponding fine-tuning dialog(s).

To obtain contextualized word representations, we used the output of the last layer of the large uncased BERT model~\cite{Devlin2019}. When using character-level tokenization, embedding representations of the characters were trained together with the network to capture relations between them. To generate the segment representations, we used 100 filters in each \ac{CNN} and aggregated the results using the max-pooling operation. Finally, the specialization layers were implemented as \acp{ReLU}~\cite{Nair2010} with 200 neurons and 50\% dropout probability~\cite{Srivastava2014}.

For transfer learning purposes, we relied on the model that achieved the top performance on the Switchboard Dialog Act Corpus in our study on dialog act recognition~\cite{Ribeiro2019b}. Thus, the weights of the \acp{CNN} and the character-level embedding layer were set to those of that model and fixed during the training phase.

%
%

\section{Results \& Discussion}
\label{sec:results}

In this section, we present and discuss the results of our experiments. We start by looking into the results achieved on segments that have communicative functions in the \textit{Task} dimension. Then, we look into the results achieved in the scenario in which the identification of the segments that have communicative functions in that dimension is also considered. Finally, we discuss the importance of the multiple components of the architecture and of transfer learning processes while looking into the results of our ablation studies.

\subsection{Task Dimension}
\label{sec:results:task}

Table~\ref{tab:results:task} shows the results of our experiments on the automatic recognition of general-purpose communicative functions in segments that have a communicative function in the \textit{Task} dimension. When evaluating using leave-one-dialog-out cross-validation, we can see that, on average, the hierarchical classifier outperformed the flat one by 3.49 percentage points in terms of exact match ratio and 2.74 in terms of hierarchical F-measure. This suggests that the per-level specialization layers are able to capture the information that is most important for distinguishing the communicative functions at a given level and that the output cascade is able to capture information regarding the hierarchical dependencies between the communicative functions at different levels.

\begin{table}
    \centering
    \begin{tabular}{l l r r r r}
        \toprule
        \textbf{Evaluation} & \textbf{Approach} & \textbf{MR} & \textbf{hP} & \textbf{hR} & \textbf{hF} \tabularnewline
        \midrule
        \multirow{2}{*}{Dialog Folds}
        & Flat         & 62.69$\pm$.17 & 82.40$\pm$.12 & 80.91$\pm$.15 & 81.65$\pm$.13 \tabularnewline
        & Hierarchical & \textbf{66.18$\pm$.47} & \textbf{85.59$\pm$.16} & \textbf{83.20$\pm$.24} & \textbf{84.39$\pm$.19} \tabularnewline
        \midrule
        \multirow{2}{*}{Corpus Folds}
        & Flat         & 46.30$\pm$.04 & 72.22$\pm$.06 & 70.14$\pm$.67 & 71.17$\pm$.38 \tabularnewline
        & Hierarchical & \textbf{47.20$\pm$.49} & \textbf{74.26$\pm$.25} & \textbf{71.53$\pm$.34} & \textbf{72.87$\pm$.27} \tabularnewline
        \bottomrule
    \end{tabular}
    \caption{Results achieved while predicting general-purpose communicative functions in the DialogBank segments that have a communicative function in the \textit{Task} dimension. The top block refers to the results achieved when evaluating using leave-one-dialog-out cross-validation, while the bottom block refers to those achieved when evaluating using leave-one-corpus-out cross-validation.}
    \label{tab:results:task}
\end{table}

When evaluating using leave-one-corpus-out cross-validation, we can observe similar patterns. However, the average exact match ratio and F-measure of the best approach decrease by 18.98 and 11.52 percentage points, respectively. In addition to the lower number of segments used to train the classifiers, part of this drop in performance is explained by the fact that, as discussed in Section~\ref{ssec:datasets}, the corpora featured in the DialogBank have different characteristics and, thus, some communicative functions mostly or only occur in one of the corpora. Consequently, the drop is slightly higher in terms of recall (11.67 percentage points) than precision (11.33 percentage points). This reveals the importance of having a representative amount of dialogs for training, whic covers all the communicative functions.

\begin{table}
    \centering
    \begin{tabular}{l r r r r r r r}
        \toprule
        \textbf{Level} & \textbf{MR} & \textbf{None\%} & \textbf{MR\textbackslash None} & \textbf{FNone} & \textbf{LC} & \textbf{NoneP} & \textbf{NoneR} \tabularnewline
        \midrule
        L1 & 87.30 &  0.00 & 87.30 &   0.00 & 12.70 &     - &      - \tabularnewline
        L2 & 80.95 &  0.00 & 80.95 &   0.00 & 19.05 &     - &      - \tabularnewline
        L3 & 80.32 &  0.00 & 80.32 &   0.00 & 19.68 &     - &      - \tabularnewline
        L4 & 70.75 & 44.90 & 58.12 &  27.76 & 14.12 & 71.69 &  86.25 \tabularnewline
        L5 & 90.16 & 90.43 & 26.17 &  65.42 &  8.41 & 93.33 &  96.93 \tabularnewline
        L6 & 99.82 & 99.82 &  0.00 & 100.00 &  0.00 & 99.82 & 100.00 \tabularnewline
        \bottomrule
    \end{tabular}
    \caption{Per-level results of one of the runs of the hierarchical approach to predict general-purpose communicative functions in the DialogBank segments that have a communicative function in the \textit{Task} dimension. \textit{None\%} refers to the percentage of segments that have no communicative function at the corresponding level. \textit{MR\textbackslash None} refers to the exact match ratio on the segments that have a communicative function at that level. \textit{FNone} refers to the percentage of those segments which were falsely classified as not having a communicative function, while \textit{LC} refers to confusion among labels. \textit{NoneP} and \textit{NoneR} refer to the precision and recall of segments that have no communicative function.}
    \label{tab:results:task-levels}
\end{table}

Overall, by looking at the different metrics, we can see that there is a gap of at least 18.21 percentage points between the results in terms of exact match ratio and hierarchical F-measure. This shows that even when the classifiers fail to predict the correct communicative function of a segment, they still predict part of the path correctly. Furthermore, the higher precision than recall suggests that the classifiers avoid predicting communicative functions that are deeper in the hierarchy. To confirm this, we randomly selected one of the runs of the hierarchical approach and looked into the full communicative function paths predicted for the segments. This way, we were able to assess the performance of the classifier on each level of the hierarchy independently, as shown in Table~\ref{tab:results:task-levels}. We can see that, when all segments are considered, the lowest exact match ratio is on Level 4, which is the one with the highest number of possible communicative functions. However, if we do not include segments that should be classified as \textit{None} in the computation of the exact match ratio, we can see that it decreases as the depth increases. Furthermore, in the levels in which there are segments misclassified as \textit{None}, this kind of misclassification is predominant in relation to confusion among the actual communicative functions of the level. On the other hand, when only considering the segments that should be classified as \textit{None}, the recall is higher than precision and increases with depth. This confirms the bias of the classifier towards the prediction of shallower functions. Overall, this can be justified by the reduction in number of labeled examples with depth, which biases the deeper output layers towards the prediction of the \textit{None} label. Still, as previously discussed, the impact of this issue can be attenuated by training the approach on set of dialogs with more extensive coverage of the communicative functions that are deeper in the hierarchy.

\subsection{All Segments}
\label{sec:results:all}

Table~\ref{tab:results:all} shows the results of our experiments which also considered the automatic recognition of the segments that have a communicative function in the \textit{Task} dimension. We can see that, in this scenario, the hierarchical approach still outperforms the flat one in terms of hierarchical F-measure by 2.24 and 1.99 percentage points when evaluating using leave-one-dialog-out and leave-one-corpus-out cross-validation, respectively. However, the performance of both approaches is similar in terms of exact match ratio. Furthermore, the wide gap between exact match ratio and hierarchical F-measure no longer exists, with a maximum difference of 2.22 percentage points. In fact, when using the flat approach, the exact match ratio surpasses the hierarchical F-measure.

\begin{table}
    \centering
    \begin{tabular}{l l r r r r}
        \toprule
        \textbf{Evaluation} & \textbf{Approach} & \textbf{MR} & \textbf{hP} & \textbf{hR} & \textbf{hF} \tabularnewline
        \midrule
        \multirow{3}{*}{Dialog Folds}
        & Flat         & 73.16$\pm$.10 & 77.57$\pm$.03 & 69.14$\pm$.58 & 73.11$\pm$.33 \tabularnewline
        & Hierarchical & 73.13$\pm$.21 & \textbf{77.64$\pm$.32} & 73.20$\pm$.31 & \textbf{75.35$\pm$.31} \tabularnewline
        & Two-Step     & \textbf{73.35$\pm$.17} & 76.84$\pm$.25 & \textbf{73.64$\pm$.15} & 75.20$\pm$.17 \tabularnewline
        \midrule
        \multirow{3}{*}{Corpus Folds}
        & Flat         & 66.14$\pm$.16 & 68.83$\pm$.10 & 62.13$\pm$.52 & 65.30$\pm$.30 \tabularnewline
        & Hierarchical & 66.12$\pm$.19 & \textbf{70.55$\pm$.23} & \textbf{64.33$\pm$.17} & \textbf{67.29$\pm$.14} \tabularnewline
        & Two-Step     & \textbf{66.82$\pm$.14} & 70.52$\pm$.20 & 63.44$\pm$.15 & 66.79$\pm$.14 \tabularnewline
        \bottomrule
    \end{tabular}
    \caption{Results achieved while predicting general-purpose communicative functions in the DialogBank segments, attributing the \textit{None} label to those which do not have a communicative function in the \textit{Task} dimension. The top block refers to the results achieved when evaluating using leave-one-dialog-out cross-validation, while the bottom block refers to those achieved when evaluating using leave-one-corpus-out cross-validation. The two-step approach applies a binary approach to decide whether a segment has a general-purpose communicative function before applying the hierarchical approach.}
    \label{tab:results:all}
\end{table}

Overall, the exact match ratio is higher than when the segments that do not have a communicative function in the \textit{Task} dimension are not considered because, in most cases, they are easy to identify. On the other hand, the F-measure is lower since misclassifying a segment as not having communicative functions in the \textit{Task} dimension means that all the functions in the correct path are missed. Additionally, since the segments without communicative functions in the \textit{Task} dimension have the \textit{None} label in every level of the hierarchy, it is expected that the classifiers become even more biased towards the prediction of shallower communicative functions.

Looking at the results, we can see that the performance drop is higher in terms of recall than precision, which is suggestive of the bias towards the prediction of shallower communicative functions. To confirm it, we also looked into the per-level results of one of the runs of the hierarchical approach. These results are shown in Table~\ref{tab:results:all-levels}. In comparison with the results in Table~\ref{tab:results:task-levels}, we can see that the exact match ratio on segments that have a communicative function on the \textit{Task} dimension is lower and the percentage of error due to misclassifications with the \textit{None} label is higher. On the other hand, the performance on the detection of segments that do not have general-purpose communicative functions at each of the levels is higher not only in terms of recall, but also in terms of precision. While the former is explained by the bias of the classifiers, the latter is explained simply by the addition of the segments with no communicative function in the \textit{Task} dimension, which account for over half of the segments in the DialogBank.

\begin{table}
    \centering
    \begin{tabular}{l r r r r r r r}
        \toprule
        \textbf{Level} & \textbf{MR} & \textbf{None\%} & \textbf{MR\textbackslash None} & \textbf{FNone} & \textbf{LC} & \textbf{NoneP} & \textbf{NoneR} \tabularnewline
        \midrule
        L0 & 85.85 & 52.63 & 83.45 &  16.55 &  0.00 & 85.52 &  88.00 \tabularnewline
        L1 & 80.97 & 52.63 & 72.36 &  17.44 & 10.20 & 84.97 &  88.73 \tabularnewline
        L2 & 78.86 & 52.63 & 67.62 &  17.80 & 14.58 & 84.74 &  88.97 \tabularnewline
        L3 & 78.77 & 52.63 & 67.35 &  17.89 & 14.76 & 84.69 &  89.05 \tabularnewline
        L4 & 80.93 & 73.90 & 41.72 &  51.30 &  6.98 & 83.95 &  94.78 \tabularnewline
        L5 & 94.87 & 95.47 & 18.69 &  74.77 &  6.54 & 96.52 &  98.49 \tabularnewline
        L6 & 99.92 & 99.92 &  0.00 & 100.00 &  0.00 & 99.92 & 100.00 \tabularnewline
        \bottomrule
    \end{tabular}
    \caption{Per-level results of one of the runs of the hierarchical approach to predict general-purpose communicative functions in all DialogBank segments. \textit{None\%} refers to the percentage of segments that have no communicative function at the corresponding level. \textit{MR\textbackslash None} refers to the exact match ratio on the segments that have a communicative function at that level. \textit{FNone} refers to the percentage of those segments which were falsely classified as not having a communicative function, while \textit{LC} refers to confusion among labels. \textit{NoneP} and \textit{NoneR} refer to the precision and recall of segments that have no communicative function.}
    \label{tab:results:all-levels}
\end{table}

As discussed in Section~\ref{sec:results:task}, we believe that the bias towards the prediction of shallower communicative functions can be attenuated by training the hierarchical approach on a sufficiently large amount of dialogs with representative coverage of the communicative functions that are deeper in the hierarchy. Still, in an attempt to attenuate the bias without relying on additional annotated data, we explored the use of a two-step approach which uses a binary classifier to identify the segments which have a communicative function in the \textit{Task} dimension before applying the hierarchical approach to those segments. 

In Table~\ref{tab:results:all}, we can see that the two-step approach achieves the highest performance in terms of exact match ratio. However, it is outperformed by the hierarchical approach in terms of hierarchical F-measure. The average performance of the binary classifier in terms of exact match ratio is of 85.52 percentage points when evaluating using leave-one-dialog-out cross validation and of 84.05 percentage points when evaluating using leave-one-corpus-out cross-validation. These values are in line with those achieved by the hierarchical approach on the top level. Thus, the differences between the two approaches can be explained by two factors. On the one hand, the hierarchical part of the two-step approach is trained solely on the segments that have communicative functions in the \textit{Task} dimension. Thus, it is less biased towards the prediction of the \textit{None} label in every level, which improves the performance in terms of exact match ratio on those segments. On the other hand, an incorrect decision of the binary classifier cannot be corrected by the predictions on the lower levels using \ac{MAP} prediction. Thus, these misclassifications at the top level have a more prominent impact on the performance of the approach in terms of hierarchical F-measure. 

When evaluating using leave-one-dialog-out cross-validation, the difference between the hierarchical and two-step approaches is of just 0.22 and 0.15 percentage points in terms of average exact match ratio and hierarchical F-measure, respectively. This suggests that the overall performance of both approaches is actually similar. Still, when evaluating using leave-one-corpus-out cross-validation, the differences between the two approaches are more noticeable, with a 0.70 percentage-point difference in favor of the two-step approach in terms of exact match ratio and a 0.50 percentage-point difference in favor of the hierarchical approach in terms of hierarchical F-measure. Thus, the selection of the best approach for the automatic recognition of general-purpose communicative functions is dependent on the metric that is most relevant for subsequent tasks.   

\subsection{Ablation Studies}
\label{sec:results:ablation}

In order to assess the importance of the multiple components of our hierarchical approach to the automatic recognition of ISO 24617-2 general-purpose communicative functions, we performed a set of ablation studies, in which one of the components was removed and the performance was compared with that of the full approach. These studies can be split into two categories: those regarding the architecture of the approach and those regarding data and transfer learning aspects. The experiments were performed in the scenario that targets the recognition of general-purpose communicative functions in segments that have a communicative function in the \textit{Task} dimension and were evaluated using leave-one-dialog-out cross-validation. Table~\ref{tab:results:ablation} shows the results of these experiments, with the first block showing the performance of the full approach for comparison.

\begin{table}
    \centering
    \begin{tabular}{l r r r r}
        \toprule
        \textbf{Approach} & \textbf{MR} & \textbf{hP} & \textbf{hR} & \textbf{hF} \tabularnewline
        \midrule
        Full                    & 66.18$\pm$.47 & 85.59$\pm$.16 & 83.20$\pm$.24 & 84.39$\pm$.19 \tabularnewline
        \midrule
        - Cascading Outputs     & 62.85$\pm$.08 & 83.79$\pm$.06 & 80.80$\pm$.07 & 82.27$\pm$.06 \tabularnewline
        - Specialization Layers & 60.58$\pm$.17 & 82.65$\pm$.10 & 79.82$\pm$.05 & 81.21$\pm$.07 \tabularnewline
        - MAP Prediction        & 63.83$\pm$.18 & 84.63$\pm$.19 & 81.82$\pm$.14 & 83.20$\pm$.17 \tabularnewline
        \midrule
        - Transfer Learning     & 59.48$\pm$.19 & 83.75$\pm$.09 & 80.35$\pm$.11 & 82.00$\pm$.09 \tabularnewline
        - LEGO Dialogs          & \textbf{70.21$\pm$.32} & \textbf{88.29$\pm$.02} & \textbf{85.24$\pm$.02} & \textbf{86.73$\pm$.01} \tabularnewline
        \bottomrule
    \end{tabular}
    \caption{Results of the ablation studies. The first block shows the performance of the full approach for comparison. The second block shows the results achieved when one component of the architecture is removed. The last block shows the results achieved without including information from external data.}
    \label{tab:results:ablation}
\end{table}

The second block of Table~\ref{tab:results:ablation} shows the results achieved when one of the components of the architecture is removed. We can see that the performance is negatively impacted in every case. By removing the connections between the output at each level and those below it, the average performance decreased by 3.33 percentage points in terms of exact match ratio and 2.12 percentage points in terms of hierarchical F-measure. Furthermore, while the performance drops in terms of both precision and recall, the drop is higher in terms of the latter. This suggests that the cascading outputs help attenuating the bias towards the prediction of shallower communicative functions.

By removing the specialization layers, the impact on performance is higher than that of removing the cascading outputs. More specifically, the average performance drops 5.60, 2.94, 3.38, and 3.18 percentage points in terms of exact match ratio and hierarchical precision, recall, and F-measure, respectively. This means that the specialization layers are able to capture the information present in the representation of the segment decorated with context information that is most relevant to distinguish among the communicative functions at the corresponding level of the hierarchy. 

Finally, by replacing the \ac{MAP} prediction approach with a top-down iterative one that limits the choices at each level of the hierarchy by only considering the communicative functions that are children of that selected for the level above it, the performance decreased 2.35 percentage points in terms of exact match ratio and 1.19 in terms of F-measure. This confirms that the \ac{MAP} prediction approach can attenuate the impact of misclassifications in the top levels by relying on the distributional outputs obtained using \textit{softmax} to obtain a joint prediction for all levels of the hierarchy.

The last block in Table~\ref{tab:results:ablation} shows the results achieved without pre-training the segment representation layers for dialog act recognition on the Switchboard Dialog Act Corpus and without considering the dialogs of the LEGO corpus during the training phase. First of all, we can see that using transfer learning improves the performance in terms of every metric, especially exact match ratio, which improves by 6.70 percentage points. This shows that, given the reduced amount of data, training the layers that generate the segment representations solely on the dialogs annotated according to the ISO 24617-2 standard leads to overfitting. On the other hand, since the Switchboard Dialog Act Corpus is sufficiently large, a model trained on its dialogs generates representations that capture information regarding generic intention that is not specific to a single set of labels. The specificities of different sets are then captured by the specialization and output layers, leading to the generation of models that have higher generalization potential.

On the other hand, the performance increased when the LEGO dialogs were not considered during the training phase, leading to an average exact match ratio of 70.21\% and an average hierarchical F-measure of 86.73\%. This is contrary to what we first observed when studying the mapping of the original dialog labels of the LEGO corpus into the communicative functions defined by the ISO 24617-2 standard and the use of those dialogs to help in the recognition of the communicative functions in the DialogBank~\cite{Ribeiro2020}. This can be explained by several factors. First, on that study, we did not rely on BERT word embeddings nor on transfer learning processes to obtain segment representations. Thus, in that case, the LEGO dialogs were important for generating more generic segment representations and avoiding overfitting. On the other hand, when those dialogs only contribute for the training of the specialization and output layers as in this study, they have a negative impact because the LEGO dialogs have different characteristics than those in DialogBank, especially in terms of the user utterances. Second, considering that the hierarchical approach has one pair of specialization and output layers per level, it has more trainable parameters than the flat one. Consequently, it is more prone to overfitting. That is relevant in this context since the LEGO dialogs are in larger number, they are highly repetitive, and are mostly covered by a small set of general-purpose communicative functions. Last, we were not focusing on the generalization ability of the classifiers in that study. Thus, the cross-validation process fine-tuned the classifiers to achieve the highest performance on the test dialog of the corresponding fold. Given this fine-tuning of the classifiers, they were less prone to overfit to the specificities of the LEGO dialogs. 

%
%

\section{Conclusions}
\label{sec:conclusion}

In this article, we have explored the automatic recognition of the general-purpose communicative functions defined by the ISO 24617-2 standard for dialog act annotation. To do so, we proposed modifications to existing approaches to flat dialog act recognition that allow them to deal with the hierarchical classification problem posed by these communicative functions. Experiments on the DialogBank, which is a reference set of dialogs annotated according to the standard, have shown that our hierarchical approach outperforms a flat approach similar to those used on most dialog act recognition tasks, both in terms of exact match ratio and hierarchical F-measure.

Addressing the modifications more specifically, instead of a single output layer, our approach includes one output layer per level of the hierarchy. This allows it to focus on distinguishing among communicative functions that are at the same level without having to deal with the ambiguities caused by communicative functions that are ancestors or descendants of each other. Furthermore, it also includes one specialization layer per level, which captures the information provided by the generic segment representation decorated with context information that is most relevant for the corresponding level.

Since the segments in a dialog may have a communicative function that is not a leaf of the hierarchy, we included an additional label in each level of the hierarchy, which refers to the absence of label at that level and those under it, allowing the prediction of paths with variable length. Furthermore, in order to avoid predicting invalid communicative function paths, our approach relies on a prediction approach based on \ac{MAP} estimation, which considers the parent-child relations between the communicative functions and improves robustness to misclassifications in individual levels.

Finally, since the DialogBank only features a small set of dialogs, we relied on transfer learning processes to generate more generic segment representations. More specifically, the layers that generate the representations were pre-trained on the largest corpus annotated for dialog acts. The generic representations are then fine-tuned for the distinction among communicative functions by the specialization layers. This way, the classifier is less prone to overfit to the characteristics of specific DialogBank dialogs, leading to improved performance. We also explored the use additional training dialogs, annotated using label mapping processes. However, they harmed performance when paired with the pre-trained segment representation layers. This happened because the specialization layers became overfit to the characteristics of the additional dialogs, which are in larger number than those in the DialogBank and deal with a specific domain.

To the best of our knowledge, this was the first study to focus on the automatic recognition of the complete hierarchy of general-purpose communicative functions defined by the ISO 24617-2 standard. Still, it focused on devising an approach that is appropriate to deal with the hierarchical classification problem posed by the communicative functions. Thus, as future work, it would be interesting to compare the different segment and context information representation approaches used in dialog act recognition studies to identify the most appropriate for this task. Furthermore, in addition to the general-purpose communicative functions, the ISO 24617-2 standard also defines dimension-specific communicative functions and a complete dialog act annotation includes additional information. Thus, the automatic recognition of all the relevant aspects should also be addressed as future work. However, that requires a representative amount of annotated data, which the DialogBank does not possess. Consequently, additional efforts should be made to increase the number of publicly available dialogs fully annotated according to the standard. 

%

%

\section*{Acknowledgements}

Eugénio Ribeiro is supported by a PhD scholarship granted by \ac{FCT}, with reference SFRH/BD/148142/2019. Additionally, this work was supported by Portuguese national funds through \ac{FCT}, with reference UIDB/50021/2020.

%

\vskip 0.2in
\bibliography{references}
\bibliographystyle{theapa}

\end{document}